\documentclass[]{interact}

\usepackage{color}
\usepackage[dvipsnames]{xcolor}

\usepackage{epstopdf}
\usepackage[caption=false]{subfig}

\usepackage[numbers,sort&compress]{natbib}
\bibpunct[, ]{[}{]}{,}{n}{,}{,}
\renewcommand\bibfont{\fontsize{10}{12}\selectfont}
\makeatletter
\def\NAT@def@citea{\def\@citea{\NAT@separator}}
\makeatother

\theoremstyle{plain}

\theoremstyle{definition}

\theoremstyle{remark}

\usepackage{multirow}
\usepackage{graphicx}
\usepackage{tabularx}

\usepackage{url} 

\begin{document}

\title{
    DM$^2$RM: Dual-Mode Multimodal Ranking for Target Objects and\\
    Receptacles Based on Open-Vocabulary Instructions
}

\author{
    \name{
        Ryosuke Korekata,
        \thanks{
            CONTACT Ryosuke Korekata. Email: rkorekata@keio.jp
            \\\\This is a preprint of an article submitted for consideration in ADVANCED ROBOTICS, copyright Taylor \& Francis and Robotics Society of Japan; ADVANCED ROBOTICS is available online at http://www.tandfonline.com/.
        }
        Kanta Kaneda, Shunya Nagashima, Yuto Imai, Komei Sugiura
    }
    \affil{
        Keio University, 3-14-1 Hiyoshi, Kohoku, Yokohama, Kanagawa 223-8522, Japan
    }
}

\maketitle

\begin{abstract}
In this study, we aim to develop a domestic service robot (DSR) that, guided by open-vocabulary instructions, can carry everyday objects to the specified pieces of furniture.
Few existing methods handle mobile manipulation tasks with open-vocabulary instructions in the image retrieval setting, and most do not identify both the target objects and the receptacles.
We propose the Dual-Mode Multimodal Ranking model (DM$^2$RM), which enables images of both the target objects and receptacles to be retrieved using a single model based on multimodal foundation models.
We introduce a switching mechanism that leverages a mode token and phrase identification via a large language model to switch the embedding space based on the prediction target.
To evaluate the DM$^2$RM, we construct a novel dataset including real-world images collected from hundreds of building-scale environments and crowd-sourced instructions with referring expressions.
The evaluation results show that the proposed DM$^2$RM outperforms previous approaches in terms of standard metrics in image retrieval settings.
Furthermore, we demonstrate the application of the DM$^2$RM on a standardized real-world DSR platform including fetch-and-carry actions, where it achieves a task success rate of 82\% despite the zero-shot transfer setting.
Demonstration videos, code, and more materials are available at https://kkrr10.github.io/dm2rm/.
\end{abstract}


\section{Introduction
}

In today’s aging society, the shortage of caregivers at home has become a serious problem.
A promising solution to this problem is the use of domestic service robots (DSRs) to physically assist care recipients~\cite{yamamoto2019development}.
Although natural language interfaces are user-friendly, the ability of DSRs to comprehend the instructions given by humans regarding household tasks (e.g., fetch-and-carry) remains insufficient.

\begin{figure*}[t]
    \centering
    \includegraphics[width=\linewidth]{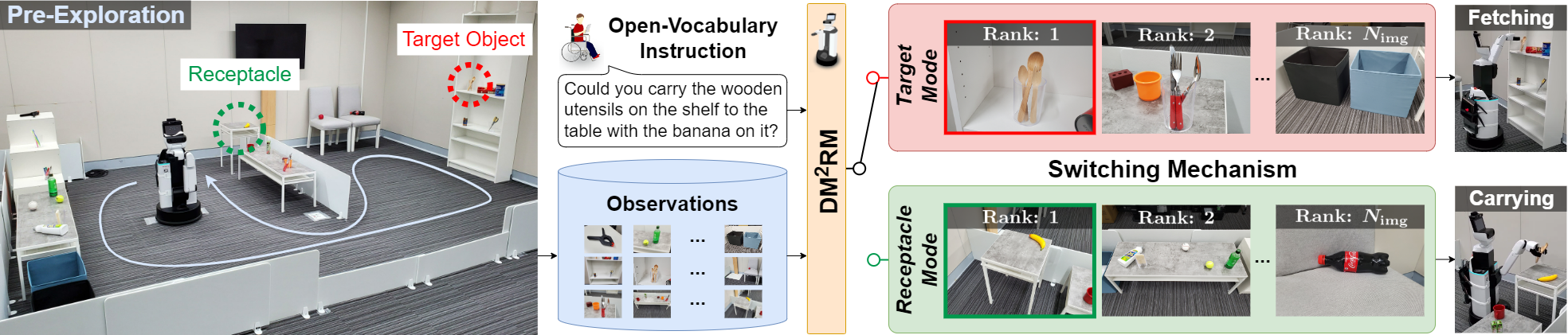}
    \caption{Overview of our method. First, the DSR collects images of the environment through pre-exploration. Given the open-vocabulary instruction, it is required to retrieve the red and green framed images as the target object image and receptacle image from the collected images, respectively. Subsequently, the DSR carries the target object to the receptacle, based on the user-selected images.}
    \label{fig:eye_catch}
\end{figure*}

In this study, we aim to develop a DSR system that, guided by open-vocabulary instructions, can carry everyday objects to specified pieces of furniture by retrieving images of the target objects and receptacles from the collected images of an environment.
Fig.~\ref{fig:eye_catch} shows an overview of our method.
First, the DSR collects images of the indoor environment through pre-exploration.
Next, an instruction such as ``Could you carry the wooden utensils on the shelf to the table with the banana on it?'' is given to the DSR.
In this case, the target object and the receptacle are `the wooden utensils on the shelf' and `the table with the banana on it,' respectively.
The DSR is required to retrieve the target object image and receptacle image from the collected images.
Subsequently, the DSR should carry the wooden utensils to the table on which the banana is placed, based on the target object image and the receptacle image selected by the users.
In this framework, it is crucial to rank these specific images higher than irrelevant images, because presenting a limited number of images can reduce the cognitive load on the users.

It is challenging for DSRs to identify the target object or the receptacle in the environment because open-vocabulary instructions given by humans are often complex and/or ambiguous.
Furthermore, if either the target object or the receptacle is misidentified, the entire task is considered unsuccessful.
In a recent open-vocabulary mobile manipulation competition~\cite{homerobotovmmchallenge2023}, the overall success rate of the winning team was just 10.8\%~\cite{melnik2023uniteam}.

Fetch-and-carry tasks based on user instructions, which are closely related to our task, have been widely studied~\cite{iocchi2015robocup,okada2019competitions,yenamandra2023homerobot}.
However, few existing methods handle mobile manipulation tasks with open-vocabulary instructions in the image retrieval setting (e.g., \cite{kaneda2024learning}).
Moreover, most such methods do not identify both the target objects and the receptacles.
Applying these methods simply to our task would be inefficient and achieve insufficient performance because of the need to train separate models specialized for the target objects and receptacles.

In this study, we propose the Dual-Mode Multimodal Ranking model (DM$^2$RM), a novel method that enables the retrieval of images of both the target objects and receptacles using a single model.
Unlike existing methods, the DM$^2$RM switches between \textit{target mode} and \textit{receptacle mode} using a single model.
This is achieved by employing a switching mechanism that leverages multimodal foundation models~\cite{radford2021learning,kirillov2023segment}.
By utilizing a mode token and phrase identification using a large language model (LLM) to switch the embedding space according to the prediction target, the DM$^2$RM enhances the similarity between the instruction and the correct image in each mode.

Please see our project page at this URL\footnote{https://kkrr10.github.io/dm2rm/} for code, dataset, and videos demonstrating the DM$^2$RM on a standardized real-world DSR platform.
The main contributions of this study are as follows:
\begin{itemize}
    \item We propose the DM$^2$RM, a novel method that individually retrieves images of both target objects and receptacles using a single model.
    \item We introduce the Switching Phrase Encoder (SPE) module, which employs a mode token and phrase identification through an LLM to switch the embedding space based on the prediction target.
    \item To handle open-vocabulary and redundant instructions, we introduce the Task Paraphraser (TP) module, designed to paraphrase the input instructions into a standardized format suitable for fetch-and-carry tasks.
    \item We introduce the Segment Anything Region Encoder (SARE) module, which enhances visual features regarding the shape and contour of objects by utilizing images overlaid with segmentation masks obtained by SAM~\cite{kirillov2023segment}.
\end{itemize}

\section{
    Related Work
}

\subsection{Language-Guided Embodied AI}

There have been many studies in the field of embodied AI, which combines robotics, computer vision, and natural language processing~\cite{liu2024aligning,duan2022survey}.
For example, several benchmark competitions have been conducted in which DSRs must execute fetch-and-carry tasks in standardized real-world environments, following user instructions~\cite{iocchi2015robocup,okada2019competitions,yenamandra2023homerobot}.
Although these tasks are closely related to our task, we do not use template-based instructions, but instead allow free-form open-vocabulary instructions with referring expressions.

Vision-and-language navigation (VLN~\cite{anderson2018vision}) is a representative embodied AI task involving natural language instructions.
For VLN tasks, most standard datasets~\cite{qi2020reverie,zhu2021soon} use images of real-world reconstructions from the Matterport3D (MP3D) dataset~\cite{chang2017matterport,anderson2018vision}.
However, MP3D lacks environmental diversity with only a few tens of discrete environments.
In contrast, the Habitat-Matterport 3D (HM3D) dataset~\cite{ramakrishnan2021hm3d,yadav2023habitat} provides hundreds of building-scale continuous environments.
Representative methods for retrieving images of target objects from images obtained through pre-exploration have been successfully applied not only in VLN (e.g., \cite{sigurdsson2023rrex}) but also in mobile manipulation tasks (e.g., \cite{chen2023open}).
Unlike these methods, the proposed DM$^2$RM employs a more practical setting that allows users to select the correct images from the top-$K$ retrieved images.

Recently, many studies have considered the application of foundation models such as LLMs and vision-language models to robotic tasks~\cite{hu2023toward,firoozi2023foundation,kawaharazuka2024realworld}.
Most existing methods utilize LLMs for commonsense reasoning~\cite{wu2023tidybot,Driess2023palme}, hierarchical planning~\cite{ichter2023can,song2023llm,hazra2024saycanplay}, or code generation~\cite{singh2023progprompt,liang2023code}.
Unlike these existing methods, our approach leverages an LLM for the switching mechanism in the SPE module, which conditions the model by identifying relevant phrases from instructions (see Section~\ref{spe}).

\subsection{Multimodal Language Understanding}

There have been numerous studies in the field of multimodal language understanding~\cite{uppal2022multimodal,chen2023vlp}.
In this subsection, we focus on referring expression comprehension (REC), object manipulation instruction understanding, and image retrieval.

REC tasks involve grounding the target object in a single image based on a single referring expression (e.g., \cite{yu2016modeling}).
However, our focus is on identifying a set of target objects and receptacles from multiple images in an environment.
Thus, most existing methods for REC tasks are not directly applicable to our task.

Most existing methods for understanding object manipulation instructions identify the target objects with bounding boxes~\cite{hatori2018interactively,korekata2023switching} or segmentation masks (e.g., \cite{iioka2023multimodal}) specified by referring expressions.
Image retrieval settings that provide multiple candidates are relatively practical, and such methods based on template-based~\cite{guadarrama2014open,nguyen2020robot} or open-vocabulary instructions (e.g., \cite{kaneda2024learning}) have been proposed.
In \cite{kaneda2024learning}, a method that handles the learning-to-rank physical objects (LTRPO) task is introduced.
LTRPO is similar to our task, but does not consider referring expressions regarding receptacles.

The standard datasets for image retrieval tasks take only text (e.g., \cite{young2014image}) or a pair of images and the associated modification text~\cite{liu2021image,han2017automatic} as input.
For image retrieval tasks, large-scale vision-and-language pre-trained models (e.g., CLIP~\cite{radford2021learning}, \cite{chen2020uniter}) have recently achieved performance improvements in the zero-shot transfer setting.
However, most methods have not been designed for inputs containing complex referring expressions.
To address this problem, we introduce the TP module to paraphrase instructions suitable for fetch-and-carry tasks and the SPE module to obtain fine-grained text features (see Sections~\ref{tp} and \ref{spe}).

\section{Problem Statement
}

In this paper, we define the Image Retrieval-based Open-Vocabulary Fetch-and-Carry (IROV-FC) task as follows: given an open-vocabulary instruction for a fetch-and-carry task from a user, the DSR retrieves images of the target object and the receptacle and subsequently transports the target object to the designated location.
This task comprises two sub-tasks: image retrieval and action execution.
In the image retrieval phase, it is desirable for the images of the target object and the receptacle to be ranked highly in their respective output ranked lists.
In the action execution phase, the DSR is expected to grasp the target object and carry it to the receptacle.
Note that the target object and the receptacle are identified from each user-selected image.

Fig.~\ref{fig:eye_catch} shows a typical scene of the IROV-FC task.
First, the DSR collects images of the indoor environment through pre-exploration.
Given the instruction ``Could you carry the wooden utensils on the shelf to the table with the banana on it?,'' the DSR is required to retrieve from the set of collected images the red and green framed images in Fig.~\ref{fig:eye_catch} as the target object image and receptacle image, respectively.
The DSR subsequently carries the wooden utensils to the table on which the banana is placed, based on the target object image and the receptacle image selected by the user.

The input and output of the IROV-FC task are defined as follows:
\begin{itemize}
    \item \textbf{Input:} an instruction and images taken in an indoor environment.
    \item \textbf{Output:} two image lists ranked based on the target object and receptacle, respectively.
\end{itemize}
The terminology used in this paper is defined as follows:
\begin{itemize}
    \item \textbf{Instruction:} an open-vocabulary instruction for a fetch-and-carry task.
    \item \textbf{Target object:} an everyday object identified as the target in the instruction.
    \item \textbf{Target object image:} an image containing the target object.
    \item \textbf{Receptacle:} a piece of furniture identified as the designated placement location in the instruction.
    \item \textbf{Receptacle image:} an image containing the receptacle.
\end{itemize}

In this study, we assume that images of the indoor environment have already been collected through pre-exploration.
This is a realistic setting because DSRs are typically used in the same indoor environment for long periods of time.
It is also assumed that trajectory generation regarding navigation, object grasping, and object placement is based on heuristic methods (see Section \ref{imple}).

\section{Proposed Method
}

\begin{figure*}[t]
    \centering
    \includegraphics[width=0.96\linewidth]{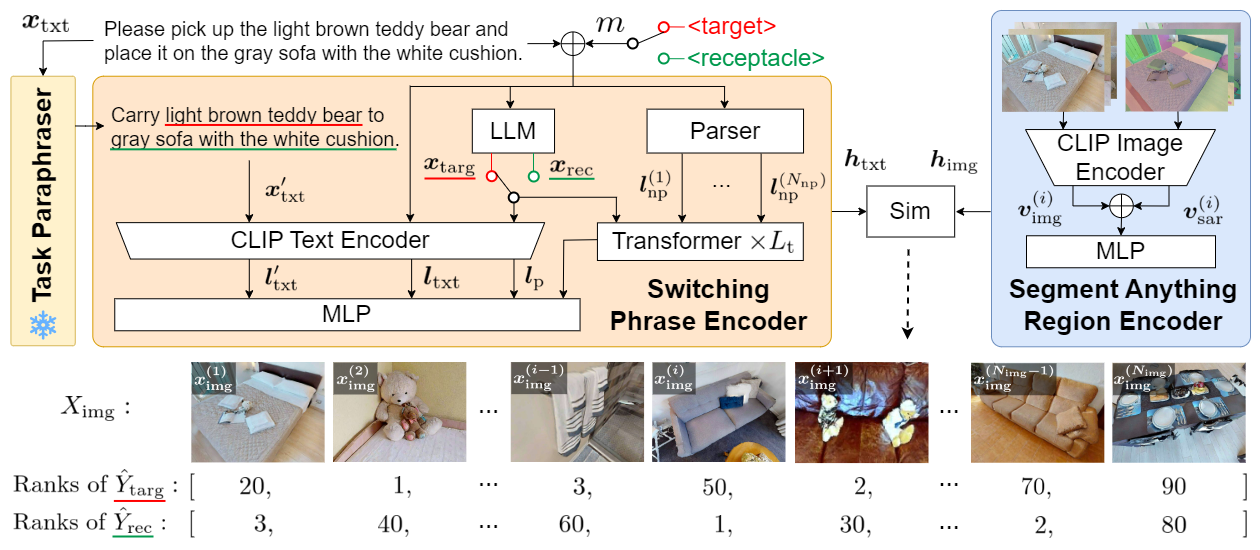}
    \caption{Architecture of the DM$^2$RM. `MLP,' `Sim,' and `$\oplus$' represent the multi-layer perceptron, cosine similarity, and concatenation, respectively.}
    \label{fig:model}
\end{figure*}


Fig.~\ref{fig:model} shows the structure of the proposed method, which mainly consists of three modules: Task Paraphraser (TP), Switching Phrase Encoder (SPE), and Segment Anything Region Encoder (SARE).
The proposed method is closely related to fetch-and-carry tasks with natural language instructions.
In these tasks, the DSR carries the target object to the receptacle following user instructions~\cite{iocchi2015robocup,okada2019competitions,yenamandra2023homerobot}.
We focus on a setting in which the users select the target object or receptacle from the presented image lists, which are ranked according to open-vocabulary instructions.

In this study, we employ the SPE module to handle object manipulation instructions with a set of target objects and receptacles.
Our approach is broadly applicable to multimodal language comprehension tasks involving, for example, a single target object, a single receptacle, and multiple sets of target objects and receptacles.
The novelties of our method are as follows:
\begin{itemize}
    \item The proposed DM$^2$RM is a novel approach that retrieves images of both target objects and receptacles individually using a single model.
    \item We introduce the SPE module, which leverages a mode token and phrase identification via an LLM to switch the embedding space according to the prediction target.
    \item To handle open-vocabulary and redundant instructions, we introduce the TP module, which paraphrases the input instructions into a standardized format suitable for fetch-and-carry tasks.
    \item We introduce the SARE module, which utilizes images overlaid with segmentation masks obtained by SAM~\cite{kirillov2023segment} to enhance visual features regarding the shape and contour of objects.
\end{itemize}

\subsection{Input}

The input $\bm{x}$ to our model is defined as follows:
\begin{align*}
    \bm{x} &= \left\{m, \bm{x}_\mathrm{txt}, X_\mathrm{img}\right\},\\
    X_\mathrm{img} &= \left\{\bm{x}_\mathrm{img}^{(i)}\right\}_{i=1}^{N_\mathrm{img}},
\end{align*}
where $m \in \{\langle \mathrm{target} \rangle, \langle \mathrm{receptacle} \rangle\}$, $\bm{x}_\mathrm{txt} \in \{0, 1\}^{V \times L}$, and $\bm{x}_\mathrm{img}^{(i)} \in \mathbb{R}^{3 \times W \times H}$ denote the mode token indicating the basis for the ranking, a tokenized instruction, and an image with width $W$ and height $H$, respectively.
Here, $V$, $L$, and $N_\mathrm{img}$ denote the vocabulary size, maximum token length, and number of images to be ranked, respectively.

\subsection{
    Task Paraphraser
    \label{tp}
}

The TP module paraphrases $\bm{x}_\mathrm{txt}$ into the standardized format $\bm{x}^{\prime}_\mathrm{txt}$ suitable for the IROV-FC task using an LLM (GPT-3.5).
Open-vocabulary instructions sometimes include redundancy or grammatical errors, making it unclear which phrases should be focused on.
This module enables such instructions to be handled in a unified manner.

We obtain $\bm{x}^{\prime}_\mathrm{txt}$ by identifying the phrases related to the target object and receptacle using an LLM.
For instance, when $\bm{x}_\mathrm{txt}$ is ``Could you, if you \textit{does} not mind, \textit{to} pick up the cardboard box and move it over towards the couch next to the fireplace?,'' the TP module outputs $\bm{x}^{\prime}_\mathrm{txt}$ as ``Carry the cardboard box to the couch next to the fireplace.''
Note that $\bm{x}^{\prime}_\mathrm{txt}$ is used as the auxiliary input to the SPE module, as explained in Section \ref{spe}.
Moreover, this module is expected to be effective for other household tasks (e.g., open/close) because of the flexible design of standard formats.

\subsection{
    Switching Phrase Encoder
    \label{spe}
}

The SPE module switches the embedding space of text features according to $m$.
In the IROV-FC task, it is necessary to predict both the target object and receptacle from a single instruction.
However, it is inefficient to train separate models specialized for various prediction tasks.

To solve this problem, we adopt a switching mechanism that enables training and inference using a single model.
Our method has two modes, \textit{target mode} and \textit{receptacle mode}, determined by $m$.
In the \textit{target mode}, the target object image is expected to be ranked highly, whereas in the \textit{receptacle mode}, the receptacle image should be ranked highly.
Note that $X_\mathrm{img}$ is the same regardless of mode.

The input to the module consists of $m$, $\bm{x}_\mathrm{txt}$, and $\bm{x}^{\prime}_\mathrm{txt}$.
First, we concatenate $m$ at the head of $\bm{x}_\mathrm{txt}$ to condition the model.
Similar to the TP module, an LLM is used to identify the phrases, $\bm{x}_\mathrm{targ}$ and $\bm{x}_\mathrm{rec}$, regarding the target object and receptacle, respectively.
To avoid focusing on irrelevant expressions, we select either of them as $\bm{x}_\mathrm{p}$, depending on the mode.
Next, noun phrases $\{\bm{x}_\mathrm{np}^{(i)}\}_{i=1}^{N_\mathrm{np}}$ are extracted from $\bm{x}_\mathrm{txt}$ using a parser~\cite{schuster2016enhanced} to obtain fine-grained text features from instructions containing multiple referring expressions.
Here, $N_\mathrm{np}$ denotes the maximum number of noun phrases.
We obtain text features $\bm{l}_\mathrm{txt} \in \mathbb{R}^{d_\mathrm{ct}}$, $\bm{l}^{\prime}_\mathrm{txt} \in \mathbb{R}^{d_\mathrm{ct}}$, $\bm{l}_\mathrm{p} \in \mathbb{R}^{d_\mathrm{ct}}$, and $\{\bm{l}_\mathrm{np}^{(i)} \in \mathbb{R}^{d_\mathrm{ct}}\}_{i=1}^{N_\mathrm{np}}$ from $\bm{x}_\mathrm{txt}$, $\bm{x}^{\prime}_\mathrm{txt}$, $\bm{x}^{\prime}_\mathrm{p}$, and $\{\bm{x}_\mathrm{np}^{(i)}\}_{i=1}^{N_\mathrm{np}}$, respectively, using the pretrained CLIP text encoder~\cite{radford2021learning}.
Here, $d_\mathrm{ct}$ denotes the output dimension.
Finally, the output $\bm{h}_\mathrm{txt} \in \mathbb{R}^{d_\mathrm{txt}}$ is obtained as follows:
\begin{align*}
    \bm{h}_\mathrm{txt} = \mathrm{MLP}\left(\left[\bm{l}_\mathrm{p}; \bm{l}_\mathrm{txt}; \bm{l}^{\prime}_\mathrm{txt}; \mathrm{Transformer}\left(\left[\bm{l}_\mathrm{p}; \bm{l}_\mathrm{np}^{(1)}; \ldots; \bm{l}_\mathrm{np}^{(N_\mathrm{np})}\right]\right)\right]\right),
\end{align*}
where $d_\mathrm{txt}$, $\mathrm{MLP}(\cdot)$, and $\mathrm{Transformer}(\cdot)$ denote the output dimension, a multi-layer perceptron, and transformer encoder~\cite{vaswani2017attention}, respectively.

\subsection{Segment Anything Region Encoder}

In the SARE module, the visual features of $X_\mathrm{img}$ and images overlaid with segmentation masks are obtained in parallel using foundation models.
Most existing methods that simply extract features from the entire image sometimes misrecognize objects with similar colors or textures.
Therefore, we introduce auxiliary images related to the segmentation masks to enhance the visual features related to the shape and contour of objects.

The input to this module is $X_\mathrm{img}$ and the outputs are the visual features $\bm{h}_\mathrm{img} \in \mathbb{R}^{d_\mathrm{img}}$ for each image $\bm{x}_\mathrm{img}^{(i)}$.
Here, $d_\mathrm{img}$ denotes the output dimension.
First, $\bm{x}_\mathrm{sar}^{(i)} \in \mathbb{R}^{3 \times W \times H}$ are obtained by overlaying the segmentation masks derived from SAM on $\bm{x}_\mathrm{img}^{(i)}$.
We obtain visual features $\bm{v}_{\mathrm{img}}^{(i)} \in \mathbb{R}^{d_\mathrm{ci}}$ and $\bm{v}_{\mathrm{sar}}^{(i)} \in \mathbb{R}^{d_\mathrm{ci}}$ from $\bm{x}_\mathrm{img}^{(i)}$ and $\bm{x}_\mathrm{sar}^{(i)}$, respectively, using the pre-trained CLIP image encoder (ViT-L/14).
Here, $d_\mathrm{ci}$ denotes the output dimension.
These features are concatenated and input to the multi-layer perceptron to obtain $\bm{h}_\mathrm{img}$.
Finally, the similarity score between $\bm{h}_\mathrm{txt}$ and $\bm{h}_\mathrm{img}$ is calculated as follows:
\begin{align*}
    \mathrm{sim}\left(\bm{x}_\mathrm{txt}, \bm{x}_\mathrm{img}^{(i)}\right) =\
    \frac{\
        \bm{h}_\mathrm{txt} \cdot \bm{h}_\mathrm{img}\
    }{\
        \| \bm{h}_\mathrm{txt} \|  \| \bm{h}_\mathrm{img} \|\
    }.
\end{align*}
The output is the ranked list of $X_\mathrm{img}$ arranged in descending order based on $\mathrm{sim}(\bm{x}_\mathrm{txt}, \bm{x}_\mathrm{img}^{(i)})$.
Two image lists, $\hat{Y}_\mathrm{targ}$ and $\hat{Y}_\mathrm{rec}$, are obtained through a total of two inferences, with $m = \langle \mathrm{target} \rangle$ and $m = \langle \mathrm{receptacle} \rangle$ specified in the input, respectively.

We use the loss function for each batch $\mathcal{L}_\mathcal{B}$ as follows:
\begin{align*}
    \mathcal{L}_\mathcal{B} = - \frac{1}{|\mathcal{B}|} \sum_{\bm{x}_\mathrm{img}^{(i)} \in \mathcal{B}}
    \log{\
    \frac{\
        \exp{ \left(\mathrm{sim}\left( \bm{x}_\mathrm{txt}, \bm{x}_\mathrm{img}^{(i)} \right) \right)}\
    }{\
        \sum_{\bm{x}_\mathrm{img}^{(j)} \in \mathcal{B}}\
        \exp{\left(\mathrm{sim}\left( \bm{x}_\mathrm{txt}, \bm{x}_\mathrm{img}^{(j)} \right)\right)} }\
    },
\end{align*}
where $|\mathcal{B}|$ denotes the batch size.
$\mathcal{L}_\mathcal{B}$ is equivalent to the scenario where only $\bm{x}_\mathrm{txt}$ is considered in InfoNCE~\cite{oord2018representation}.

\section{
    Experiments
}

\subsection{
    Dataset
}

We built the novel Learning-To-Rank in Real Indoor Environments for Fetch-and-Carry (LTRRIE-FC) dataset for the IROV-FC task.
The LTRRIE-FC dataset is based on the HM3D~\cite{ramakrishnan2021hm3d,yadav2023habitat} and MP3D~\cite{chang2017matterport,anderson2018vision} datasets.
To the best of our knowledge, there is no standard dataset for the IROV-FC task.
The standard datasets for VLN tasks (e.g., \cite{qi2020reverie}) and LTRPO tasks (e.g., \cite{kaneda2024learning}) are not suitable for IROV-FC because they do not consider the task of transporting the target object to the receptacle.
Furthermore, most existing datasets were constructed from the MP3D dataset, which lacks environmental diversity as it only includes a few tens of environments.
In contrast, HM3D is a large-scale dataset containing hundreds of building-scale environments of 3D real-world reconstructions.
However, there is no standard dataset that contains natural language instructions annotated by humans for the HM3D dataset.
Therefore, we annotated instructions for images collected from both the HM3D and MP3D datasets.

\begin{figure}[t]
    \centering
    \includegraphics[width=\linewidth]{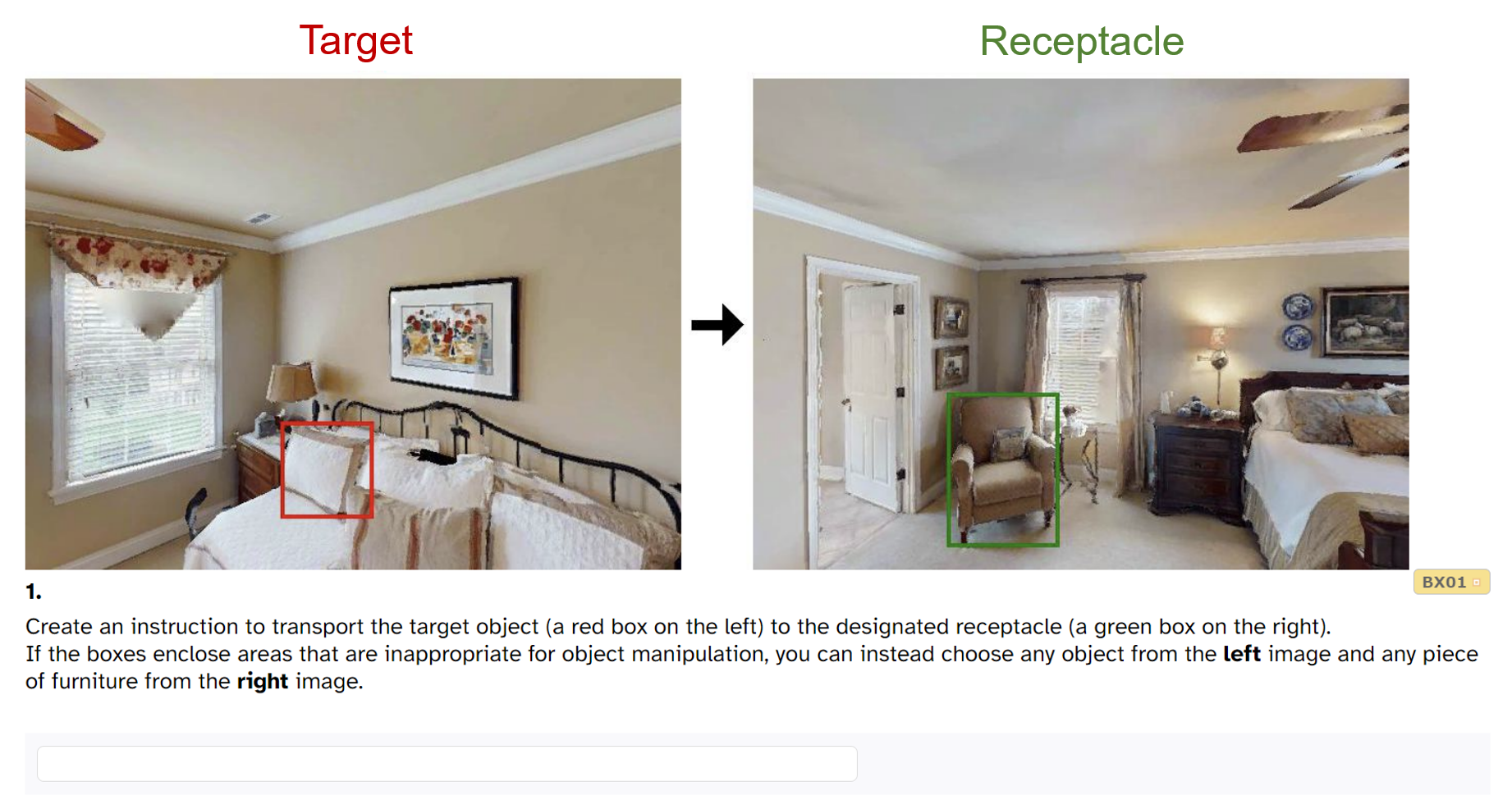}
    \caption{Annotation interface. Annotators were required to give instructions for the DSR to carry the target object (a red bounding box) to the receptacle (a green bounding box). These instructions were input in the text box below the images.}
    \label{fig:anno}
\end{figure}

To collect the images from the continuous environments in HM3D, we used \cite{savva2019habitat} to simulate the exploration of the environments by a DSR.
The map of each environment was provided by HM3D.
The DSR captured images at randomly selected viewpoints defined on grid points.
At each viewpoint, the camera pose was set towards the nearest object or piece of furniture.
The DSR also captured images of the surroundings by rotating the camera pose 60\textdegree\ to the left and right at a fixed height.
These steps were conducted on each floor in the environment.
The procedure for collecting images from MP3D followed that described in \cite{kaneda2024learning}.

In the LTRRIE-FC dataset, a sample consists of an instruction, a target object image, and a receptacle image.
To extract target object images and receptacle images from the collected images, we used Detic~\cite{zhou2022detecting}, an open-vocabulary object detector, as follows:
First, we defined 121 and 48 target object classes (e.g., `pillow,' `book,' `cup') and receptacle classes (e.g., `shelf,' `table,' `bed'), respectively.
These categories were selected from the classes listed in \cite{yadav2023habitat}.
Next, Detic was applied to each image.
The images in which the target object class was detected were used as target object images.
We selected receptacle images using the same procedure.
For data cleansing, we manually removed samples for which the detected bounding box was extremely small, the detected object did not fit within the image, or there were significant mesh reconstruction artifacts.
Finally, a target object image and a receptacle image in the same environment were combined to create a sample.

The instructions in the LTRRIE-FC dataset were collected by 226 annotators using the SoSci Survey\footnote{https://www.soscisurvey.de/} service.
Fig.~\ref{fig:anno} shows the annotation interface.
The annotators were presented with two images, one of a target object and one of a receptacle.
They were then asked to give instructions for transporting the target object (in the red bounding box) to the receptacle (in the green bounding box) (e.g., ``Pick up the white cushion on the sofa and place it on the brown armchair near the bed.'').
However, the depicted bounding boxes sometimes enclosed inappropriate areas for object manipulation tasks due to misdetection (e.g., an ungraspable object such as a window was detected as the target object).
In such cases, the annotators were allowed to select a more appropriate object from the left image and a piece of furniture from the right image instead.
Data from annotators that repeatedly input the same instruction or had short response times were excluded to improve the quality of the dataset.

The LTRRIE-FC dataset consists of 6,581 English instructions and 7,148 images collected from 774 real-world indoor environments.
It has a vocabulary size of 2,491, a total of 103,263 words, and an average sentence length of 15.69 words.
The LTRRIE-FC dataset includes 5,814, 354, and 413 samples in the training, validation, and test sets, respectively.
These sets contain 690, 42, and 42 environments, respectively, without duplication of environments.
Therefore, the objects in the test sets can be regarded as unseen.
We built two subsets for the test set, HM3D-FC and MP3D-FC, depending on the type of environment from which the samples were obtained.
We used the training set to train the model, the validation set to tune the hyperparameters, and the test sets to evaluate the model.

\subsection{Parameter Settings}


\begin{table}[t]
    \normalsize
    \centering
    \caption{Experimental settings of DM$^2$RM. Here, \#$L_\mathrm{t}$, \#$H$, and \#$A$ denote the number of layers, hidden size, and attention heads of the transformer encoder in the SPE module, respectively.}
    \label{tab:params}
    \begin{tabular}{lc}
        \toprule
        Optimizer & Adam ($\beta_1=0.9$, $\beta_2=0.98$) \\
        Learning rate & $1\times10^{-4}$ \\
        Batch size & $128$ \\
        \#Epoch & $20$ \\
        Dropout & $0.4$ \\
        Transformer & \#$L_\mathrm{t}$: $5$, \#$H$: $768$, \#$A$: $4$ \\
        \bottomrule
        \end{tabular}
\end{table}

Table~\ref{tab:params} shows the experimental settings of the proposed method.
Our model had approximately 71M trainable parameters and 309G multiply-add operations.
We trained our model on a GeForce RTX 3090 with 24 GB of GPU memory and an Intel Core i9-10900KF with 64 GB of RAM.
Training our model for $20$ epochs took approximately 1 h.
The inference time for computing the similarity between a single instruction and a single image was approximately 14.8 ms.

During every epoch, we measured the mean reciprocal rank (MRR) and recall@10 of the model on the validation set.
The final performance on the test sets were based on the model achieving the maximum sum of recall@10 and MRR on the validation set.

\subsection{Quantitative Results}

\begin{table*}[t]
    \centering
    \normalsize
    \caption{Quantitative comparison between the DM$^2$RM and baseline methods on the HM3D-FC test set. The best score for each metric is in \textbf{bold}. $^*$ denotes reproduced results.}
    \label{tab:quant_dataset_hm3d}
        \begin{tabular}{@{\hspace{1mm}}l@{\hspace{1.8mm}}lc@{\hspace{0.3mm}}cc@{\hspace{2mm}}c@{\hspace{2mm}}c@{\hspace{2mm}}c}
            \toprule
            \multicolumn{1}{c}{\multirow{2}{*}{[\%]}} & \multirow{2}{*}{Method} & \multicolumn{2}{c}{Prediction} & \multicolumn{4}{c}{HM3D-FC (unseen)} \\
            \cmidrule(lr){3-4} \cmidrule(lr){5-8}
            & & \textit{Targ.} & \textit{Rec.} & MRR$\uparrow$ & R@5$\uparrow$ & R@10$\uparrow$ & R@20$\uparrow$ \\
            \midrule
            (i) & CLIP~\cite{radford2021learning} & $\checkmark$ & $\checkmark$ & $10.8$ & $13.7$ & $24.9$ & $49.5$ \\
            (ii) & NLMap$^*$~\cite{chen2023open} & $\checkmark$ & $\checkmark$ & $11.8$ & $14.1$ & $26.1$ & $48.4$ \\
            (iii-a) & \multirow{2}{*}{MultiRankIt~\cite{kaneda2024learning}} & $\checkmark$ & & $20.5$ \small{$\pm$ $2.3$} & $30.1$ \small{$\pm$ $3.4$} & $48.2$ \small{$\pm$ $1.4$} & $73.2$ \small{$\pm$ $2.8$} \\
            (iii-b) & & & $\checkmark$ & $19.8$ \small{$\pm$ $1.1$} & $27.1$ \small{$\pm$ $3.2$} & $49.1$ \small{$\pm$ $5.9$} & $74.6$ \small{$\pm$ $3.1$} \\
            \midrule
            (iv) & DM$^2$RM (ours) & $\checkmark$ & $\checkmark$ & $\mathbf{32.0}$ \small{$\pm$ $0.5$} & $\mathbf{47.7}$ \small{$\pm$ $1.4$} & $\mathbf{67.9}$ \small{$\pm$ $0.8$} & $\mathbf{87.3}$ \small{$\pm$ $1.1$} \\ 
            \bottomrule
        \end{tabular}
\end{table*}

\begin{table*}[t]
    \centering
    \normalsize
    \caption{Quantitative comparison between the DM$^2$RM and baseline methods on the MP3D-FC test set. The best score for each metric is in \textbf{bold}. $^*$ denotes reproduced results.}
    \label{tab:quant_dataset_mp3d}
        \begin{tabular}{@{\hspace{1mm}}l@{\hspace{1.8mm}}lc@{\hspace{0.3mm}}cc@{\hspace{2mm}}c@{\hspace{2mm}}c@{\hspace{2mm}}c}
            \toprule
            \multicolumn{1}{c}{\multirow{2}{*}{[\%]}} & \multirow{2}{*}{Method} & \multicolumn{2}{c}{Prediction} & \multicolumn{4}{c}{MP3D-FC (unseen)} \\
            \cmidrule(lr){3-4} \cmidrule(lr){5-8}
            & & \textit{Targ.} & \textit{Rec.} & MRR$\uparrow$ & R@5$\uparrow$ & R@10$\uparrow$ & R@20$\uparrow$ \\
            \midrule
            (i) & CLIP~\cite{radford2021learning} & $\checkmark$ & $\checkmark$ & $15.0$ & $14.6$ & $28.5$ & $59.9$ \\
            (ii) & NLMap$^*$~\cite{chen2023open} & $\checkmark$ & $\checkmark$ & $11.5$ & $14.3$ & $25.7$ & $52.3$ \\
            (iii-a) & \multirow{2}{*}{MultiRankIt~\cite{kaneda2024learning}} & $\checkmark$ & & $26.7$ \small{$\pm$ $2.4$} & $35.9$ \small{$\pm$ $4.0$} & $52.8$ \small{$\pm$ $5.3$} & $71.1$ \small{$\pm$ $2.7$} \\
            (iii-b) & & & $\checkmark$ & $16.4$ \small{$\pm$ $1.6$} & $23.3$ \small{$\pm$ $2.0$} & $39.7$ \small{$\pm$ $5.3$} & $60.1$ \small{$\pm$ $3.7$} \\
            \midrule
            (iv) & DM$^2$RM (ours) & $\checkmark$ & $\checkmark$ & $\mathbf{36.8}$ \small{$\pm$ $1.5$} & $\mathbf{46.5}$ \small{$\pm$ $2.8$} & $\mathbf{63.5}$ \small{$\pm$ $2.8$} & $\mathbf{76.3}$ \small{$\pm$ $1.5$} \\ 
            \bottomrule
        \end{tabular}
\end{table*}

Tables~\ref{tab:quant_dataset_hm3d} and \ref{tab:quant_dataset_mp3d} show the quantitative results.
The performance of the proposed method is compared with that of several baseline methods on the HM3D-FC and MP3D-FC test sets.
The table presents the average and standard deviation over five trials.
The `Prediction' column indicates whether the model handled target objects and/or receptacles.

We used CLIP~\cite{radford2021learning}, NLMap~\cite{chen2023open}, and MultiRankIt~\cite{kaneda2024learning} as the baseline methods.
We selected CLIP because it has been successfully applied to image retrieval tasks without fine-tuning.
NLMap was selected due to its similarity to the proposed method, as it also employs a CLIP-based approach for object retrieval from images collected during pre-exploration.
The scores shown for CLIP and NLMap were obtained from a single trial because the use of the pre-trained frozen model provides consistent results across multiple trials.
MultiRankIt was selected as a baseline method because of its effective application in the LTRPO task, which is related to the IROV-FC task.
We trained two separate MultiRankIt models for target objects and receptacles, because MultiRankIt cannot output ranked lists for both target objects and receptacles with a single model.

We used MRR and recall@$K$ as evaluation metrics, with MRR as the primary metric.
This is because they are standard metrics in image retrieval settings~\cite{liu2009learning}.
MRR is defined as follows:
\begin{align*}
    \mathrm{MRR} = \frac{1}{N_\mathrm{txt}}\
    \sum_{i=1}^{N_\mathrm{txt}} \frac{1}{r_1^{(i)}},
\end{align*}
where $N_\mathrm{txt}$ and $r_1^{(i)}$ denote the number of instructions and the highest rank among the relevant images, respectively.
Recall@$K$ is defined as follows:
\begin{align*}
    \mathrm{Recall@}K =\
    \frac{1}{N_\mathrm{txt}}\
    \sum_{i=1}^{N_\mathrm{txt}} \frac{|A_i \cap B_i|}{|A_i|},
\end{align*}
where $A_i$ and $B_i$ denote the set of relevant images to be retrieved and the top-$K$ retrieved images, respectively.

Table~\ref{tab:quant_dataset_hm3d} indicates that the proposed method (iv) achieved the MRR of 32.0\%, whereas baseline methods (i), (ii), (iii-a), and (iii-b) achieved the MRR of 10.8\%, 11.8\%, 20.5\%, and 19.8\%, respectively, for the HM3D-FC test set.
Furthermore, Table~\ref{tab:quant_dataset_mp3d} indicates that the proposed method (iv) and baseline methods (i), (ii), (iii-a), and (iii-b) achieved the MRR of 36.8\%, 15.0\%, 11.5\%, 26.7\%, and 16.4\%, respectively, for the MP3D-FC test set.
Therefore, the proposed method outperformed the best baseline methods by 11.5 points and 10.1 points in terms of MRR on the HM3D-FC test set and the MP3D-FC test set, respectively.
Similarly, the proposed method outperformed the baseline methods in terms of recall@$K$ on the test sets.
The differences in performance between our method and the baseline methods were statistically significant in terms of all evaluation metrics ($p$-value $<$ 0.01).

\subsection{Qualitative Results}

\begin{figure*}[t]
    \centering
    \includegraphics[width=\linewidth]{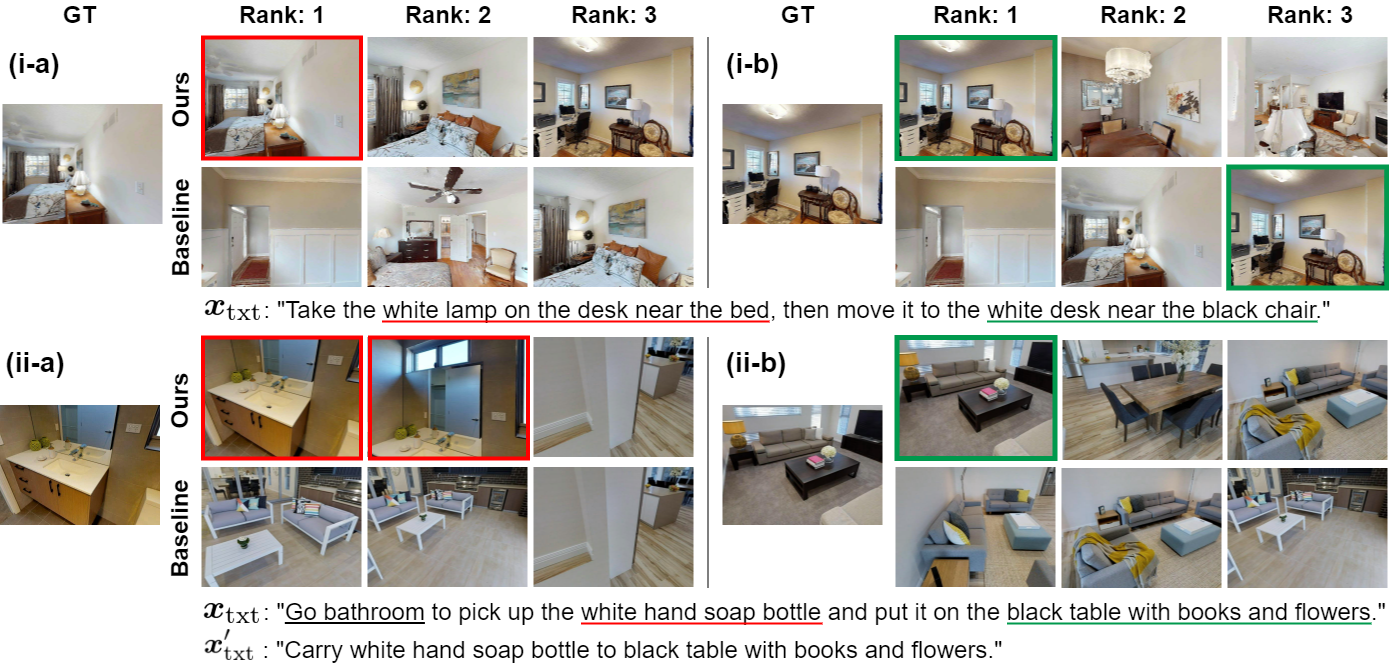}
    \caption{Qualitative comparison between our method and a baseline method~\cite{kaneda2024learning}. For each sample, $\bm{x}_\mathrm{txt}$ and/or $\bm{x}^{\prime}_\mathrm{txt}$, the top-3 retrieved images, and the GT image are shown. The results regarding the target object and receptacle are shown on the left (*-a) and right (*-b), respectively. The target object images and receptacle images are highlighted in the red and green frames, respectively. The words underlined in red, green, and black indicate $\bm{x}_\mathrm{targ}$, $\bm{x}_\mathrm{rec}$, and grammatical errors, respectively.}
    \label{fig:qual_success}
\end{figure*}

\begin{figure}[t!]
    \centering
    \includegraphics[width=\linewidth]{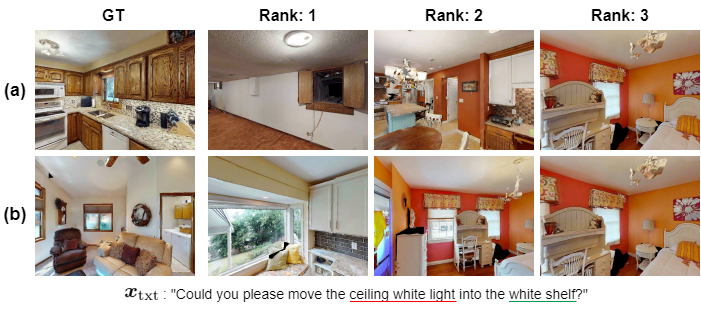}
    \caption{A failure sample on the HM3D-FC test set. Rows (a) and (b) show the qualitative results in the target and receptacle modes, respectively. From left to right: GT images and top-3 retrieved images. The words highlighted in red and green indicate $\bm{x}_\mathrm{targ}$ and $\bm{x}_\mathrm{rec}$, respectively.}
    \label{fig:qual_failure}
\end{figure}

Fig.~\ref{fig:qual_success} shows the qualitative comparison between the proposed method and one of the baseline methods~\cite{kaneda2024learning}.
The ground truth (GT) image and the top-3 retrieved images are shown for each mode.
Fig.~\ref{fig:qual_success} (i-a) and (i-b) show a sample from the HM3D-FC test set, where $\bm{x}_\mathrm{txt}$ was ``Take the white lamp on the desk near the bed, then move it to the white desk near the black chair.''
In the proposed method, $\bm{x}_\mathrm{targ}$ and $\bm{x}_\mathrm{rec}$ were `white lamp on the desk near the bed' and `white desk near the black chair,' respectively.
For this sample, the MRR of our method was 100\%, whereas that of the baseline method was 30\%.
The baseline method has incorrectly retrieved the same irrelevant image as the top-1 result in both Fig.~\ref{fig:qual_success} (i-a) and (i-b).
On the contrary, the proposed method has successfully retrieved the correct image as the top-1 for each mode.
This indicates that the switching mechanism in the SPE module works effectively.

Similarly, Fig.~\ref{fig:qual_success} (ii-a) and (ii-b) show a sample from the MP3D-FC test set, where $\bm{x}_\mathrm{txt}$ was ``\textit{Go bathroom} to pick up the white hand soap bottle and put it on the black table with books and flowers.''
In the proposed method, $\bm{x}_\mathrm{targ}$ and $\bm{x}_\mathrm{rec}$ were `white hand soap bottle' and `black table with books and flowers,' respectively.
For this sample, the proposed method and the baseline method achieved the MRR of 100\% and 20\%, respectively.
In Fig.~\ref{fig:qual_success} (ii-a), the baseline method has mistakenly retrieved images of tables, influenced by the word regarding the receptacle, whereas our method correctly has retrieved images of hand soap as ranks 1 and 2.
In Fig.~\ref{fig:qual_success} (ii-b), the proposed method appropriately handles the referring expressions regarding the color of the receptacle and its surrounding objects, whereas the baseline method does not.
These results indicate that the introduction of LLM-based phrase identification in the SPE module enhanced the similarity between the instruction and the correct image in each mode.
In addition, we believe that obtaining $\bm{x}^{\prime}_\mathrm{txt}$ using the TP module was beneficial to the proper handling of $\bm{x}_\mathrm{txt}$ with grammatical errors (e.g., ``Go bathroom \ldots'' should be ``Go to the bathroom \ldots'').

Fig.~\ref{fig:qual_failure} shows a failure sample of the proposed method on the HM3D-FC test set.
For this sample, $\bm{x}_\mathrm{txt}$ was ``Could you please move the ceiling white light into the white shelf?''
In each mode, the top-3 retrieved images did not match the GT image, resulting in the MRR of 5\%.
However, these images are not technically incorrect because they contain a white ceiling light and a white shelf in the \textit{target mode} and the \textit{receptacle mode}, respectively, as required by the instruction.
Therefore, it is hypothesized that the failure was caused by ambiguous instructions containing insufficient referring expressions.

\subsection{Ablation Studies}

\begin{table*}[t]
    \centering
    \normalsize
    \caption{Ablation studies of the DM$^2$RM on the HM3D-FC test set. The best score for each metric is in \textbf{bold}.}
    \label{tab:quant_ablation_hm3d}
        \begin{tabular}{llcccc}
            \toprule
            \multicolumn{1}{c}{\multirow{2}{*}{}} & \multirow{2}{*}{Model} & \multicolumn{4}{c}{HM3D-FC (unseen)} \\
            \cmidrule(lr){3-6}
            & & MRR$\uparrow$ [\%] & R@5$\uparrow$ [\%] & R@10$\uparrow$ [\%] & R@20$\uparrow$ [\%] \\
            \midrule
            (a) & DM$^2$RM (full) & $\mathbf{32.0}$ $\pm$ $0.5$ & $\mathbf{47.7}$ $\pm$ $1.4$ & $\mathbf{67.9}$ $\pm$ $0.8$ & $\mathbf{87.3}$ $\pm$ $1.1$ \\
            \midrule
            (b) & w/o SPE & $22.5$ $\pm$ $1.4$ & $33.2$ $\pm$ $1.8$ & $53.0$ $\pm$ $2.5$ & $78.9$ $\pm$ $2.7$ \\ 
            (c) & w/o TP & $28.4$ $\pm$ $1.4$ & $44.7$ $\pm$ $2.0$ & $66.3$ $\pm$ $0.7$ & $85.2$ $\pm$ $1.1$ \\ 
            (d) & w/o SARE & $29.7$ $\pm$ $0.6$ & $45.0$ $\pm$ $2.7$ & $64.9$ $\pm$ $1.4$ & $86.6$ $\pm$ $1.4$ \\ 
            \bottomrule
        \end{tabular}
\end{table*}

\begin{table*}[t]
    \centering
    \normalsize
    \caption{Ablation studies of the DM$^2$RM on the MP3D-FC test set. The best score for each metric is in \textbf{bold}.}
    \label{tab:quant_ablation_mp3d}
        \begin{tabular}{llcccc}
            \toprule
            \multicolumn{1}{c}{\multirow{2}{*}{}} & \multirow{2}{*}{Model} & \multicolumn{4}{c}{MP3D-FC (unseen)} \\
            \cmidrule(lr){3-6}
            & & MRR$\uparrow$ [\%] & R@5$\uparrow$ [\%] & R@10$\uparrow$ [\%] & R@20$\uparrow$ [\%] \\
            \midrule
            (a) & DM$^2$RM (full) & $\mathbf{36.8}$ $\pm$ $1.5$ & $\mathbf{46.5}$ $\pm$ $2.8$ & $\mathbf{63.5}$ $\pm$ $2.8$ & $\mathbf{76.3}$ $\pm$ $1.5$ \\
            \midrule
            (b) & w/o SPE & $21.3$ $\pm$ $0.9$ & $25.6$ $\pm$ $1.8$ & $42.0$ $\pm$ $1.7$ & $63.2$ $\pm$ $1.0$ \\ 
            (c) & w/o TP & $31.4$ $\pm$ $2.2$ & $40.5$ $\pm$ $2.6$ & $56.8$ $\pm$ $3.8$ & $75.0$ $\pm$ $0.7$ \\ 
            (d) & w/o SARE & $33.2$ $\pm$ $1.2$ & $43.4$ $\pm$ $1.6$ & $60.0$ $\pm$ $2.5$ & $75.0$ $\pm$ $2.0$ \\
            \bottomrule
        \end{tabular}
\end{table*}

Tables~\ref{tab:quant_ablation_hm3d} and \ref{tab:quant_ablation_mp3d} show the results of ablation studies.
As ablation studies, we set the following three conditions:\\
\textbf{SPE ablation:} To investigate the impact of the SPE module on performance improvement, we removed $m$ and handled both target objects and receptacles without using the switching mechanism.
Tables~\ref{tab:quant_ablation_hm3d} and \ref{tab:quant_ablation_mp3d} show that the MRR decreased by 9.5 points and 15.5 points for model (b) compared with model (a) on the HM3D-FC and MP3D-FC test sets, respectively.
This indicates that the introduction of SPE is beneficial for the IROV-FC task.\\
\textbf{TP ablation:} We removed the TP module to investigate its influence on the performance.
Tables~\ref{tab:quant_ablation_hm3d} and \ref{tab:quant_ablation_mp3d} show that there was a decrease of 3.6 points and 5.4 points in MRR for model (c) compared with model (a) on the HM3D-FC and MP3D-FC test sets, respectively.
This suggests that paraphrasing redundant instructions into a standardized format suitable for the task is an effective approach.\\
\textbf{SARE ablation:} We removed $\bm{v}_{\mathrm{sar}}^{(i)}$ on investigate the impact of the SARE module to the performance.
Tables~\ref{tab:quant_ablation_hm3d} and \ref{tab:quant_ablation_mp3d} indicate that the MRR decreased by 2.3 points and 3.6 points for model (d) compared with model (a) on the HM3D-FC and MP3D-FC test sets, respectively.
Thus, enhancing visual features associated with the shape and contour of objects is beneficial to the performance of the proposed method.

\subsection{Error Analysis}


\begin{table}[t]
    \caption{Categorization of failure cases. We selected a total of 20 samples (10 from each test set) and manually conducted a detailed error analysis on the top-5 images for each mode.}
    \label{tab:error}
    \normalsize
    \centering
    \begin{tabular}{lcc}
        \toprule
        Error Type & \textit{Target Mode} & \textit{Receptacle Mode} \\
        \midrule
        Ambiguous Instruction &
        8 & 8 \\
        Referring Expression Comprehension Error &
        7 & 2 \\
        Phrase Selection Error &
        1 & 7 \\
        Object Grounding Error &
        4 & 3 \\
        \midrule
        Total &
        20 & 20 \\
        \bottomrule
    \end{tabular}
\end{table}

We define a failure case as a sample for which the MRR fell below 10.0\%.
There were 98 failure cases (21 and 77 cases from the HM3D-FC and MP3D-FC test sets, respectively).
We selected a total of 20 samples (10 from each test set) and manually conducted a detailed error analysis on the top-5 images for each mode.

Table~\ref{tab:error} categorizes the failure cases in each mode.
The causes of failure could be divided into four types: ambiguous instruction (AI), referring expression comprehension error (RE), phrase selection error (PS), and object grounding error (OG).
The AI category refers to cases in which the retrieved images cannot be considered entirely incorrect because of some ambiguity in $\bm{x}_\mathrm{txt}$, such as the failure sample shown in Fig.~\ref{fig:qual_failure}.
The RE category refers to cases in which the category of retrieved images of objects or pieces of furniture was correct, but did not match the referring expressions contained in $\bm{x}_\mathrm{txt}$ (e.g., images containing a cushion on the red sofa were incorrectly retrieved when the target object was `the cushion on the black bed')
The PS category refers to cases in which the model retrieved images of landmark objects or pieces of furniture specified in $\bm{x}_\mathrm{txt}$ instead of the target object or receptacle (e.g., when the target object was `the desk next to the sofa,' the model mistakenly retrieved images containing only the sofa).
The OG category refers to cases in which the object grounding performance was insufficient, and so the model retrieved irrelevant images.

Table~\ref{tab:error} indicates that AI was the main bottleneck in both the target and receptacle modes.
A possible solution to these errors is to introduce multi-target settings, where a single expression can refer to an arbitrary number of objects (e.g., \cite{liu2023gres}).

\subsection{Discussion}

\begin{figure}[t]
    \centering
    \includegraphics[width=0.7\linewidth]{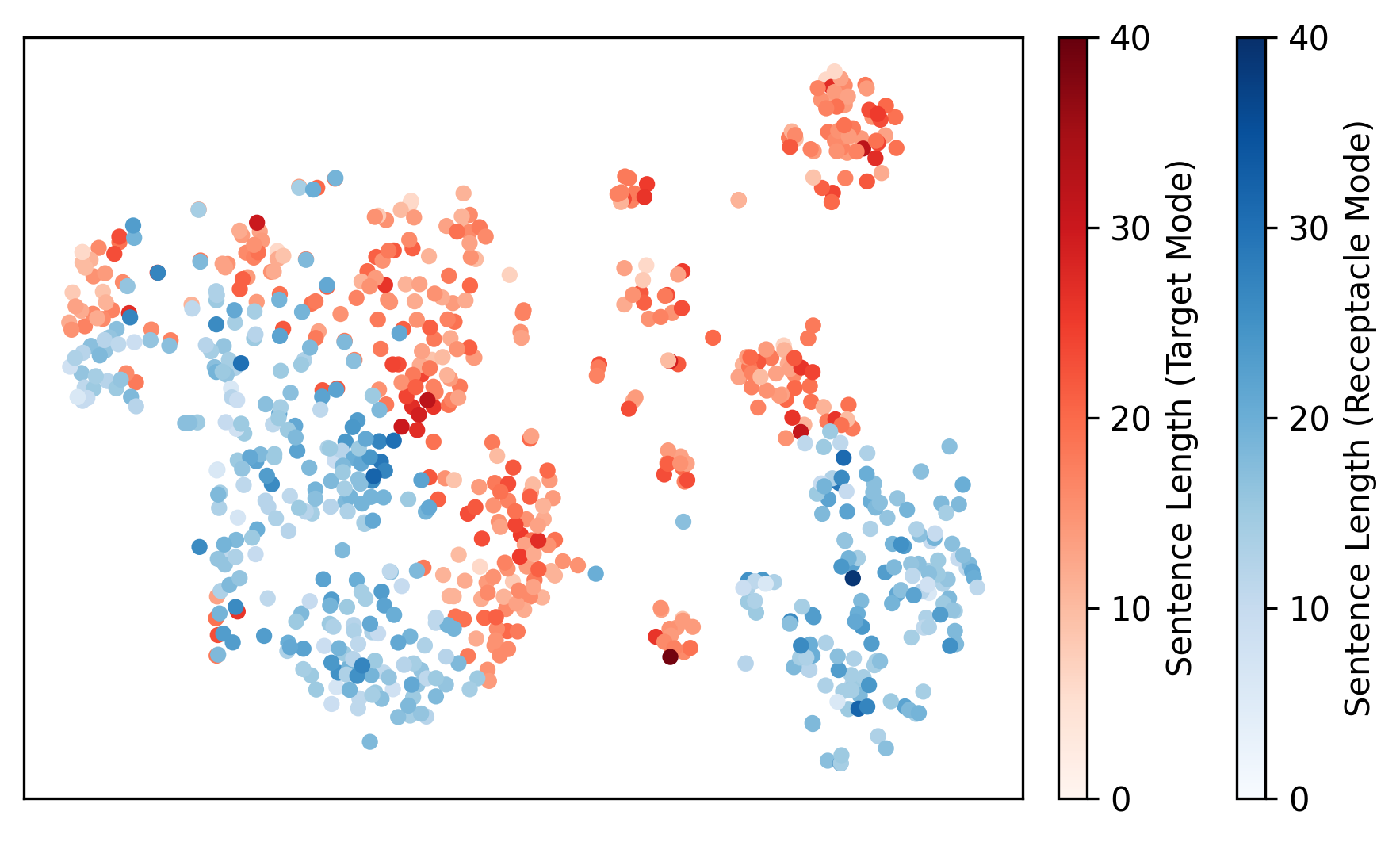}
    \caption{t-SNE~\cite{van2008visualizing} visualization of $\bm{h}_\mathrm{txt}$ for all instructions on the test sets. Each point represents the embedding space of an instruction in the \textit{target mode} (red) and the \textit{receptacle mode} (blue), with the color brightness reflecting the sentence length.}
    \label{fig:t-sne}
\end{figure}

To investigate the influence of the switching mechanism in the SPE module, we visualized the embedding space of text features when the same instructions were input with different $m$.
Fig.~\ref{fig:t-sne} shows the t-SNE~\cite{van2008visualizing} visualization of $\bm{h}_\mathrm{txt}$ for all instructions on the test sets in each mode, with the color brightness reflecting the sentence length.
The results demonstrate that the clusters of the different modes have a distinct separation, indicating that the embedding space is effectively switched according to the prediction target.
Importantly, the nonlinear distribution of the clusters implies that the switching is not purely the result of simple translations or rotations in the embedding space.

\section{
    Physical Experiments
}

We validated the proposed method in a real-world environment using a DSR.
We did not fine-tune the model by using the physical environment.
The DSR executed fetch-and-carry tasks based on the instructions given by the users.

\subsection{Settings}


\begin{figure}[t]
    \centering
        \begin{tabular}{c}
            \begin{minipage}{0.5\columnwidth}
                \centering
                \includegraphics[height=4.9cm]{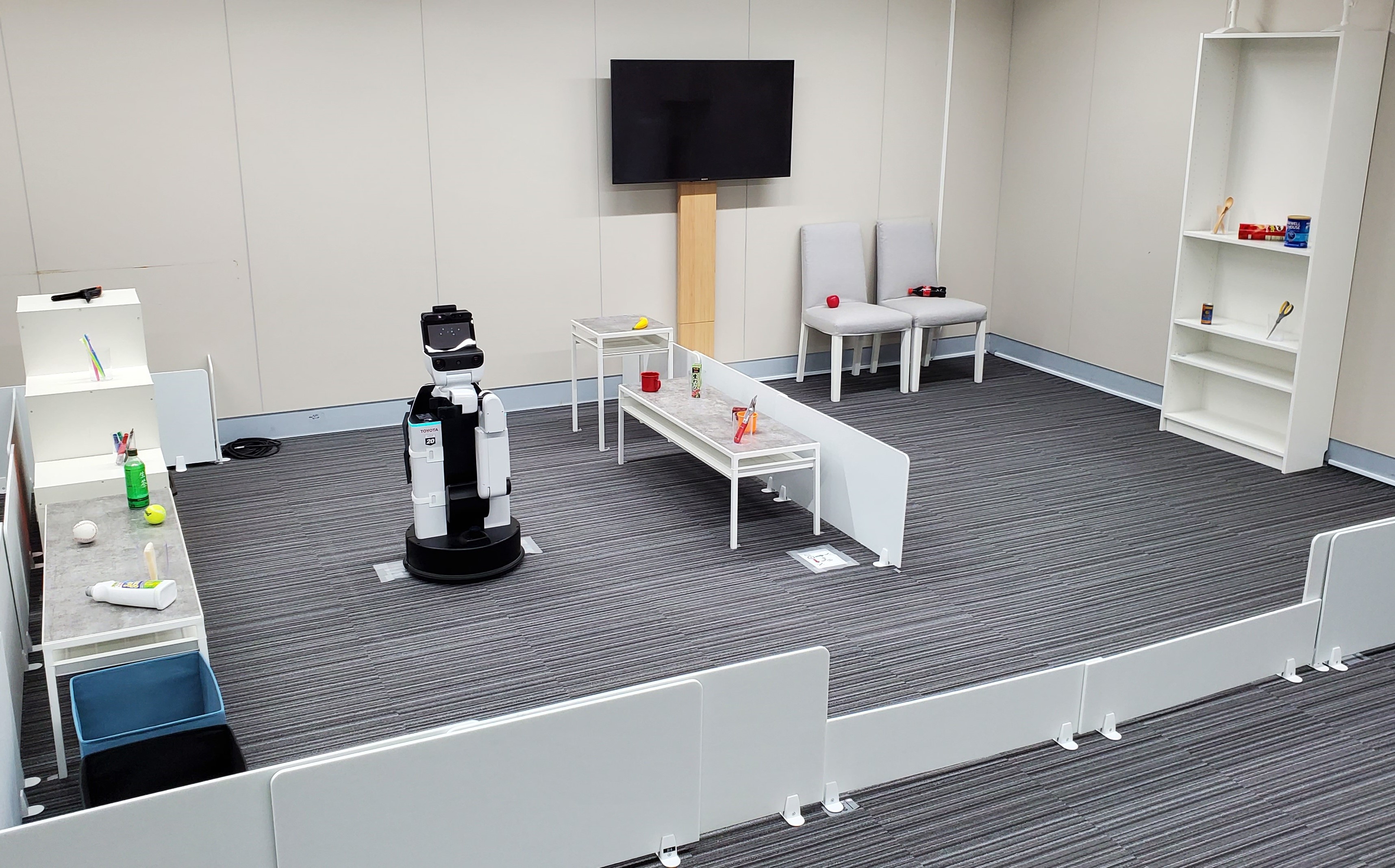} \\
                \footnotesize (i)
            \end{minipage}
            \begin{minipage}{0.5\columnwidth}
                \centering
                \includegraphics[height=4.94cm]{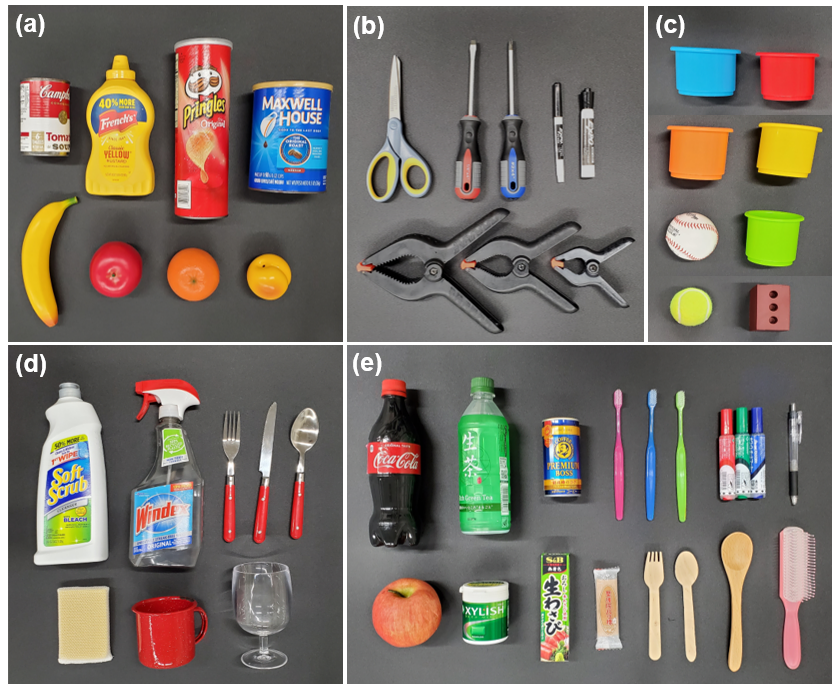} \\
                \footnotesize (ii)
            \end{minipage}
        \end{tabular}
    \caption{Experimental settings. (i) Domestic environment standardized in WRS2020RS~\cite{okada2019competitions}. (ii) Everyday objects used in the physical experiments. These objects are (a) `Food items,' (b) `Tool items,' (c) `Shape items,' and (d) `Kitchen items' from the YCB object set~\cite{calli2015benchmarking}, and (e) general unseen objects.}
    \label{fig:env_objects}
\end{figure}

Fig.~\ref{fig:env_objects} (i) shows the experimental environment.
The environment replicated the standardized environment of the World Robot Summit 2020 Partner Robot Challenge/Real Space (WRS2020RS~\cite{okada2019competitions}), which was an international contest focusing on benchmark tidy-up tasks in home environments.
The size of this environment was $6.0 \times 4.0 \; \mathrm{m}^2$, and it featured nine pieces of furniture arranged as shown in Fig.~\ref{fig:env_objects} (i).
Users randomly selected one piece of furniture as the receptacle when providing instructions to the DSR.
We used the Human Support Robot~\cite{yamamoto2019development} developed by the Toyota Motor Corporation.
This mobile manipulator has been used as the standard platform of the RoboCup@Home competition~\cite{iocchi2015robocup} since 2017.

Fig.~\ref{fig:env_objects} (ii) shows a total of 50 everyday objects used in the physical experiments.
These objects are part of the YCB object set~\cite{calli2015benchmarking}, which includes standard objects for manipulation research.
Furthermore, we included general unseen objects to enrich the diversity in terms of appearance and sizes.
We conducted experiments with 10 unique object placement patterns.
In each object placement pattern, 20--30 unseen objects selected from Fig.~\ref{fig:env_objects} (ii) were placed in random positions on randomly selected pieces of furniture.
Several small objects (e.g., toothbrush) were placed in the markerless NICT cases~\cite{magassouba2018multimodal}.

\subsection{
    Implementation
    \label{imple}
}

During the pre-exploration phase, the DSR collected images of the environment at 17 predefined viewpoints using an Asus Xtion Pro camera.
Path planning and navigation were based on standard methods using a map created in advance.

Next, the users gave English open-vocabulary instructions to the DSR.
The users were required to provide instructions with referring expressions to carry an arbitrary object (see Fig.~\ref{fig:env_objects} (ii)) to an arbitrary receptacle (see Fig.~\ref{fig:env_objects} (i)), such as ``Please pick up the sponge near the cleanser and put it in the blue box.''
For each object placement pattern, 10 instructions were given, resulting in a total of 100 trials.

The behavior of the DSR after receiving the instructions was designed as follows:
The DSR first retrieved images of the target object and the receptacle from the latest stored images and presented the respective top-10 images to the users using the WebUI.
We adopted the zero-shot transfer setting using the model trained on the LTRRIE-FC dataset to test the robustness of the proposed method towards the unseen objects.
Next, the target object image and the receptacle image were selected by the users from the presented images.
If the target object image was not included in the top-10 images in the \textit{target mode}, it was regarded as a failure and the fetching action was not conducted.
Subsequently, the DSR moved to the location at which the target object image was captured and grasped the target object.
The grasp point was determined based on the point cloud obtained from the depth image and the segmentation mask of the target object.
The segmentation mask was obtained using SAM~\cite{kirillov2023segment} by inputting the point prompt given by the users regarding the target object.
Finally, the DSR attempted to carry the target object to the receptacle only if the following conditions were met: receptacle image was within the top-10 images in the \textit{receptacle mode} and the fetching action was successful.
We did not employ a learning-based approach for trajectory generation regarding object grasping and placing because this is beyond the scope of this study.

\subsection{Quantitative Results}

\begin{table}[t]
    \caption{Quantitative results of the physical experiments. The numbers in parentheses indicate $N_\mathrm{s}$ / $N_\mathrm{a}$.}
    \label{tab:quant_physical}
    \centering
    \normalsize
    \begin{tabular}{ccccc}
        \toprule
        \multirow{2}{*}{MRR$\uparrow$ [\%]} & \multirow{2}{*}{R@10$\uparrow$ [\%]} & \multicolumn{3}{c}{SR$\uparrow$ [\%]} \\
        \cmidrule(lr){3-5}
        & & Fetching & Carrying & Overall \\
        \midrule
        39 & 96 & 92 (89 / 97) & 95 (82 / 86) & 82 (82 / 100) \\
        \bottomrule
    \end{tabular}
\end{table}

We used MRR, recall@10, and the task success rate (SR) as evaluation metrics in the physical experiments.
SR is defined as $\mathrm{SR} = \frac{N_\mathrm{s}}{N_\mathrm{a}},$ where $N_\mathrm{s}$ and $N_\mathrm{a}$ denote the number of successes and attempts, respectively.
When calculating the MRR, $\frac{1}{r_1^{(i)}}$ was considered to be 0 if $r_1^{(i)}$ was greater than 10.
This is because only the top-10 images were presented to the users, following the standard UI configuration.

Table~\ref{tab:quant_physical} shows the quantitative results of the physical experiments.
The results show that the MRR, recall@10, fetching SR, carrying SR, and overall SR were 39\%, 96\%, 92\%, 95\%, and 82\%, respectively.
Despite the zero-shot transfer setting using our model trained on the LTRRIE-FC dataset, these results indicate that the performance of the proposed method remained robust when handling unseen objects in real-world environments.
These results also indicate that our model can be successfully integrated into the DSR to perform the entire scenario, including fetching and carrying actions.

\subsection{Qualitative Results}

\begin{figure*}[t]
    \centering
    \includegraphics[width=\linewidth]{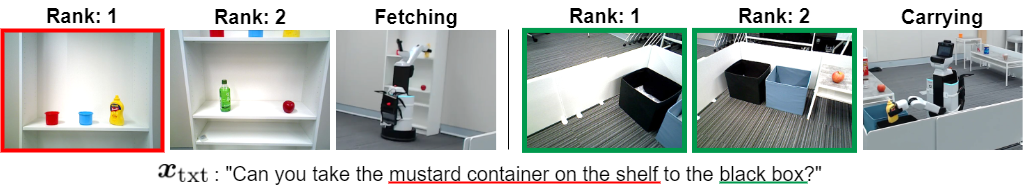}
    \caption{Qualitative results of the physical experiments. The target object images and receptacle images are framed in red and green, respectively. The words underlined in red and green indicate $\bm{x}_\mathrm{targ}$ and $\bm{x}_\mathrm{rec}$, respectively.}
    \label{fig:qual_physical}
\end{figure*}

Fig.~\ref{fig:qual_physical} shows a successful sample of the physical experiments.
For this sample, $\bm{x}_\mathrm{txt}$ was ``Can you take the mustard container on the shelf to the black box?''
The proposed method successfully retrieved the correct image as the top-1 for each mode, resulting in the MRR of 100\%.
Subsequently, the DSR grasped the mustard container and placed it in the black box.
More information along with demonstration videos and other qualitative results are available on our project page at this URL\footnote{https://kkrr10.github.io/dm2rm/}.

\section{Conclusions}

In this study, we focused on the IROV-FC task, in which a DSR retrieves images of the target object and the receptacle from stored images based on an open-vocabulary instruction, and subsequently transports the target object to the receptacle.
Our contributions are as follows:
\begin{itemize}
    \item We proposed the DM$^2$RM, a novel approach that retrieves images of both target objects and receptacles individually using a single model.
    \item We introduced the SPE module, which leverages a mode token and phrase identification via an LLM to switch the embedding space according to the prediction target.
    \item To handle open-vocabulary and redundant instructions, we introduced the TP module to paraphrase the input instructions into a standardized format suitable for fetch-and-carry tasks.
    \item We also introduced the SARE module, which utilizes images overlaid with segmentation masks obtained by SAM~\cite{kirillov2023segment} to enhance visual features regarding the shape and contour of objects.
    \item The DM$^2$RM outperformed the baseline methods in terms of the standard metrics on the LTRRIE-FC dataset, a novel dataset based on HM3D~\cite{ramakrishnan2021hm3d,yadav2023habitat} and MP3D~\cite{chang2017matterport,anderson2018vision}.
    \item In physical experiments, our method achieved a task success rate of more than 80\% in the standardized environment, despite the zero-shot transfer setting. These results indicate that the DM$^2$RM can be successfully integrated into the DSR to perform the entire scenario, including fetch-and-carry actions.
\end{itemize}

In future work, we plan to introduce multi-target settings (e.g., \cite{liu2023gres}) to handle cases in which a single expression refers to an arbitrary number of objects.

\section*{ACKNOWLEDGMENT}
This work was partially supported by JSPS KAKENHI Grant Number 23H03478, JST Moonshot, and NEDO.
\bibliographystyle{tfnlm}
\bibliography{reference}
\end{document}